\documentclass[letterpaper]{article} 
\usepackage{aaai24}  
\usepackage{times}  
\usepackage{helvet}  
\usepackage{courier}  
\usepackage[hyphens]{url}  
\usepackage{graphicx} 
\usepackage{natbib}  
\usepackage{caption} 
\usepackage{url}
\usepackage{cite}
\usepackage{amsthm,amsmath,amssymb,amsfonts}
\usepackage{array}
\graphicspath { {./figure/}}
\usepackage{textcomp}
\usepackage{xcolor}
\usepackage{algorithm2e}
\usepackage{subfigure}
\usepackage{mdframed}
\usepackage{multirow}
\usepackage{pifont}
\usepackage{booktabs}
\usepackage{float}
\usepackage{color, xcolor} 
\usepackage[toc,page]{appendix}
\usepackage[switch]{lineno}
\newtheorem{Properties}{Property}
\newtheorem{theorem}{Theorem}

\newtheorem{corollary}{Corollary}
\theoremstyle{definition}

\theoremstyle{remark}

\usepackage{newfloat}
\usepackage{listings}
\DeclareCaptionStyle{ruled}{labelfont=normalfont,labelsep=colon,strut=off} 
\floatstyle{ruled}
\newfloat{listing}{tb}{lst}{}
\floatname{listing}{Listing}
\title{PC-Conv: Unifying Homophily and Heterophily with Two-fold Filtering}
\author{
    Bingheng Li,\textsuperscript{\rm 1}
    Erlin Pan,\textsuperscript{\rm 1}
    Zhao Kang\textsuperscript{\rm 1}\footnote{Corresponding author}
}
\affiliations{
    \textsuperscript{\rm 1} University of Electronic Science and Technology of China,
    Chengdu, Sichuan, China

    \{bingheng86,wujisixsix6\}@gmail.com, zkang@uestc.edu.cn
}

\usepackage{bibentry}

\begin{document}
\maketitle

\begin{abstract}
Recently, many carefully crafted graph representation learning methods have achieved impressive performance on either strong heterophilic or homophilic graphs, but not both. Therefore, they are incapable of generalizing well across real-world graphs with different levels of homophily. This is attributed to their neglect of homophily in heterophilic graphs, and vice versa.
In this paper, we propose a two-fold filtering mechanism to extract homophily in heterophilic graphs and vice versa. In particular, we extend the graph heat equation to perform heterophilic aggregation of global information from a long distance.
The resultant filter can be exactly approximated by the Possion-Charlier (PC) polynomials. To further exploit information at multiple orders, we introduce a powerful graph convolution PC-Conv and its instantiation PCNet for the node classification task. Compared with state-of-the-art GNNs, PCNet shows competitive performance on
well-known homophilic
and heterophilic graphs. Our implementation is available at  \textit{https://github.com/uestclbh/PC-Conv}.
\end{abstract}

\section{Introduction}
Developing methods to handle graph data has received increasing attention in the past decade. Graph neural networks (GNNs), which jointly leverage topological structure and node attribute information, have achieved immense success on numerous graph-related learning tasks. For each node, GNNs recursively aggregate and transform attributes from its neighbors. Due to this information aggregation mechanism, strong homophily of the graph is an implicit condition for GNNs to achieve good performance in downstream tasks \cite{Bird}. Basically, homophily
means that similar nodes are prone to connect to each other. 
As a pioneering GNNs architecture, GCN \cite{GCN} and its variants
\cite{SGC,APPNP} all adopt a low-pass filter to achieve feature smoothing through information aggregation between neighboring nodes. We refer to these models as homophilic GNNs.

Some recent works point out that GNNs are subject to substantial performance degradation on heterophilic graphs, where connected nodes tend to have different labels \cite{Gemo,Mixhop}. 
To handle heterophilic data, many researchers try to modify the original GNN or even design new architectures and techniques \cite{H2GCN,GGCN,luan2022revisiting,ACM-GCN,Global-hete,mao2023demystifying,pan2023beyond}

Despite respectable results in heterophilic graphs, many heterophilic GNNs still struggle to maintain superiority on homophilic graphs. The over-exploitation of heterophilic nodes and carefully crafted designs often limit their performance on homophilic graphs.
In fact, homophilic and heterophilic graphs are not independent of each other, and real graphs cannot easily be classified as homophilic or heterophilic. This is because the degree of heterophily is just a statistical concept that measures the percentage of connected nodes from different classes. 
For example, in popular heterophilic datasets like Actor, Cornell, and Penn94, more than 20\% of connected nodes belong to the same class \cite{Global-hete}. This phenomenon is even more pronounced on large datasets \cite{LINKX}. Therefore, homophilic and heterophilic graphs should not be treated separately, and some graph-agnostic processing methods are desired. In particular, the homophily in heterophilic graphs should be deliberately explored, and vice versa.
Thus, this boils down to two questions:
\begin{figure*}[ht]
    \centering
    \includegraphics[width=1.0\textwidth]{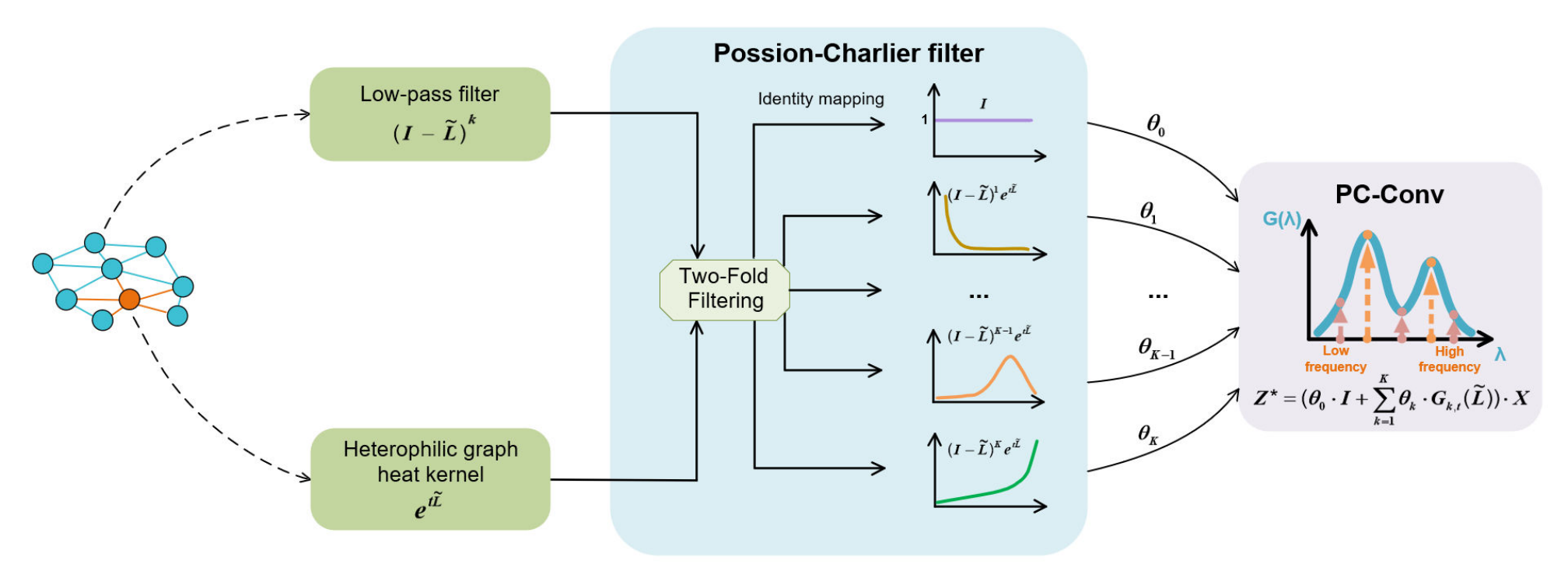}
    \caption{An illustration of our proposed PC-Conv. }
    \label{fig:scheme}
\end{figure*}

\textbf{Q1: how to deliberately design the information aggregation mechanism for homophilic and heterophilic nodes? }

\textbf{Q2: how to design a plausible strategy to combine them?}

In response to the first question, we analyze information aggregation from a local and global perspective. For homophilic nodes, the local topology helps to aggregate the neighbors and preserve the local feature information to some extent \cite{zhang2021node}. This operation can be achieved by a simple low-pass filter. For heterophilic nodes, in contrast, the local topology could diversify the characteristics of the nodes, deviating abnormally from the original features and making the nodes difficult to distinguish in downstream tasks \cite{H2GCN}. Some methods enlarge the node neighborhood to alleviate this problem. However, the size of the neighborhood is hard to set and varies from node to node. Therefore, we propose to use the global structure information of heterophilic nodes for information aggregation. Unlike the complex model in \cite{Global-hete}, we generalize the graph heat equation \cite{GDC,GraphHeat} to the heterophilic graph and precompute the heterophilic heat kernel for global information aggregation. 

For the second question, 
we propose a novel approach to combine filters based on our optimization framework: a two-fold filtering mechanism. Specifically, we integrate the heterophilic graph heat kernel with a local low-pass filter through two-fold filtering to yield a novel graph filter. To simplify its computation, we approximate it in an exact way with Possion-Charlier (PC) polynomials, dubbed as PC-filter. 

In addition, 
to capture multi-order information, we propose to use PC-filter banks with adaptive coefficients, dubbed as PC-Conv, which is shown in Fig. \ref{fig:scheme}. 
An exhaustive spectral analysis of PC-Conv is provided, including numerical approximation capability and flexibility in learning complex filters. Finally, a simple network architecture for node classification instantiates PC-Conv, i.e., PCNet, demonstrating the most advanced performance on both homophilic and heterophilic graphs.
The main contributions of this paper are summarized as follows:
\begin{itemize}
    \item \textbf{Two-fold filtering mechanism:} We develop a filtering strategy to perform both homophilic and heterophilic aggregation in any graph with different levels of homophily.
    \item \textbf{Heterophilic graph heat kernel:} 
    We extend the graph heat kernel to heterophilic graph and integrate it with a low-pass filter to form a novel PC-filter.
    \item \textbf{PC-Conv:}
    We propose PC-Conv to explore multi-order information with learnable parameters. An exhaustive spectral analysis is also presented.
    \item \textbf{PCNet:} 
    We instantiate PC-Conv with a simple GNN architecture, that is, PCNet. Experiments on node classification demonstrate the state-of-the-art (SOTA) performance.
\end{itemize}

\section{Preliminaries}\label{preliminaries}

\textbf{Notations}. 
Define an undirected graph $\mathcal{G}=(\mathcal{V},E)$ with a finite node set $|\mathcal{V}|=m$ and an edge set $E$ with $|E|$ edges. We denote $i\sim j$ if node $i$ and $j$ are adjacent, otherwise $i\not\sim j$. Let $A\in\mathbb{R}^{m \times m}$ be the adjacency matrix, $A_{i j}=\begin{cases} 1, &i\sim j\\0, &i\not\sim j\end{cases}\label{hand}$, 
whose degree matrix $D$ is diagonal with $D_{i i}=\sum_j A_{i j}$. 
Thus, the symmetric normalized adjacency matrix is $\tilde{A}=(D+I)^{-\frac{1}{2}} (A+I) (D+I)^{-\frac{1}{2}}$ and the normalized Laplacian matrix is $L=I-\tilde{A}$. We denote the node feature matrix by $X\in\mathbb{R}^{m \times d}$, where $d$ is the dimensionality of the attribute. We use $X_{i:}$ and $X_{:j}$ to denote the feature vector of node $i$ and the $j$-th graph signal, respectively.

\subsection{Spectral Graph Convolution}
Spectral graph convolution is the core of spectral GNNs \cite{ARMA} and can be unified as follows:
\begin{equation}
Z=g(L) \cdot X,   
\end{equation}
where $Z$ is the output representation matrix. The key to the spectral graph convolution is the design of graph filter $g$.  For Laplacian matrix $L=U \Lambda U^{\top}$, the eigenvalue matrix $\Lambda=diag(\lambda_{1},\dots,\lambda_{m})$ represents the frequency of the graph signal and $U=\{u_{1},\dots,u_{m}\}$ is the frequency component associated with $ \Lambda $. Given a graph signal $x$ on $\mathcal{G}$, the graph Fourier transform and the inverse transform are defined as  $\hat{x}=U^{\top} x$ and $x=U \hat{x}$, respectively. The purpose of designing a filter $ g(L)$ is to adjust the frequency response of the graph signal. The graph convolution in the spectral domain can be expressed as:
\begin{equation}
    z = g(L) \cdot x = U \cdot g(\Lambda) \cdot U^{\top} x = U \cdot g(\Lambda) \cdot\hat{x}.  
\end{equation}

\subsection{Graph Heat Equation}
\cite{GHE} defines the diffusion on the graph to characterize the global information flow, where the graph heat equation is proposed to depict the changes in node features during this process \cite{GHE-community}. 
The graph heat equation is generally written as:
\begin{equation}
\frac{d x(t)}{d t} = - L x(t)\quad \quad x(0) = x_{0},
\label{heateq}
\end{equation}
where $x(t)$ represents the feature information (graph signal) at time $t$. The solution is the heat kernel
$x(t) =e^{-t L} x_{0}$. \cite{DGC} uses the Euler method to approximate the heat kernel.  Despite these methods have achieved decent results on homophilic graphs, their current form hinders their applications on heterophilic graphs.

\subsection{Graph Optimization Framework}\label{optimization-framework}
A classic graph optimization problem is \cite{BernNet}:
\begin{equation}
\min _{Z} f(Z) = Tr(Z^{\top}h(L)Z)+\alpha \|Z-X\|_{F}^{2},
\label{gof}
\end{equation}
where the first term is the smoothness constraint on the signals based on graph topological structure \cite{p-Laplacian} and $h(\cdot)$ servers as an energy function \cite{GNN-LF-HF}. The second term enforces that $Z$ retains as much information as possible from the original feature, where $\alpha>0$ is a balance parameter. Problem (\ref{gof}) has a closed-form solution by setting $\frac{\partial f(\mathbf{Z})}{\partial \mathbf{Z}} = 0 $ :
\begin{equation}
Z = g(L) X = (\alpha I+ h(L))^{-1} X.
\label{sol4}
\end{equation}
where $h(L)$ has to be positive semi-definite. If $h (\mathbf{L})$ is not positive semi-definite, then the optimization function $f(\mathbf{Z})$ is not convex, and the solution to $\frac{\partial f(\mathbf{Z})}{\partial \mathbf{Z}} =   0$ may correspond to a saddle point. 
In terms of Eq.(\ref{sol4}), we can establish the connection between $h(\cdot)$ and $g(\cdot)$, i.e.,
\begin{equation}
h(L)=g(L)^{-1}-\alpha I.
\end{equation}
Therefore, different filters $g (L)$ correspond to different energy functions $h (L)$. 

\section{Methodology}
\subsection{Two-fold Filtering}
The operation of the spectral filter on the feature matrix is equivalent to the information aggregation in the spatial domain \cite{Analysis}. From the spatial point of view, the low-pass filter makes the connected node features smooth and converge gradually; the high-pass filter sharpens the nodal features, making them easily distinguishable. The low-pass and high-pass filters correspond to two opposite ways of aggregating information in the spatial domain, which we refer them as homophilic aggregation and heterophilic aggregation, respectively. For a real graph with different levels of homophily, a single aggregation results in information loss. Therefore, we propose a two-fold filtering mechanism that performs homophilic and heterophilic aggregation of node features, which can be deduced from graph optimization.
\begin{flushleft}
\textbf{heterophilic aggregation: }
\end{flushleft}
\begin{equation}
    Y^*  = \mathop{\arg\min}\limits_{Y} Tr(Y^{\top}h_{1}(L)Y)+ \alpha_{1} \|Y-X\|_{F}^{2} 
    \label{hete}
\end{equation}

\begin{flushleft}
\textbf{homophilic aggregation: }
\end{flushleft}
\begin{equation}
    Z^*  = \mathop{\arg\min}\limits_{Z}
Tr(Z^{\top}h_{2}(L)Z)+\alpha_{2} \|Z-Y^*\|_{F}^{2}
\label{homo}
\end{equation}
where $h_{1}(L)$, $h_{2}(L) $ represent the energy function of heterophilic aggregation and homophilic aggregation, respectively.  
We can obtain the optimal solution from Eqs. (\ref{hete}-\ref{homo}):
\begin{equation}
         Z^*=[(\alpha_{1} I+ h_{1}(L))(\alpha_{2} I+ h_{2}(L))]^{-1} X   
\end{equation}

\begin{Properties}\label{order-invariant}
 $Z^*$ is order-invariant if we exchange the heterophilic and homophilic aggregation operations. (Proof in
Appendix) 
 \end{Properties}

From another perspective, heterophilic and homophilic aggregation are conducted simultaneously. 
 In the spatial domain, two-fold filtering can be seen as pushing nodes' features away from their neighbors to make some heterophilic nodes easily distinguishable and smoothing homophilic nodes with low-order information at the same time.
\subsection{Heterophilic Graph Heat Kernel}
To address the global information aggregation challenge for heterophilic nodes, we consider it from the perspective of graph diffusion. Eq. (\ref{heateq}) mainly integrates global information in a homophilic graph. For heterophilic nodes, 
the complementary graph can provide missing-half structure, which can somehow mitigate inter-class information aggregation and enrich its structural information abundantly \cite{luan2019break,liu2021self,luan2022complete, Complementary}. We define
a complementary graph $\Bar{G}$ whose adjacency matrix $\Bar{A}=p I-A$, where $p$ is a hyperparameter to measure the self-loop information. 
Denote $\tilde{L} = (p-2)I+ L $, we can obtain the heterophilic propagation as $H_{hete}=e^{t \tilde{L}}X$. 
Based on Taylor expansion, we can formulate the information diffusion of the heterophilic graph as:  
\begin{align*}
&H_{hete}=\sum_{r=0}^{\infty} \frac{(t(p-1)I-A)^{2r+1}}{(2r+1)!}+\frac{(A-t(p-1)I)^{2r}}{(2r)!}.
\end{align*}
We can see that the heterophilic graph heat kernel pushes the odd-order neighbors away and aggregates the neighborhood information for even-order neighbors, which is consistent with the structural property of the heterophilic graph. Different from existing works that only exploit even-order neighbors \cite{EvenNet}, our different treatments of odd-order and even-order neighbors allow us to dig for more holistic information. Furthermore, the heterophilic graph heat kernel can make use of infinite order information decaying with increasing order, which is also more flexible than \cite{GCN,gat,graphsage} that fix the neighborhood size.

\subsection{PC-Conv}\label{PC-Conv}

For heterophilic aggregation, $g_{1}(\tilde{L})=e^{t\tilde{L}}$ and the corresponding energy function is $h_{1}(\tilde{L})=e^{-t\tilde{L}}-\alpha_1 I$. For homophilic aggregation, we use a low-pass filter to capture local low-order information and set $h_{2}(\tilde{L})$ to $\tilde{L}$ for simplicity. 

According to the requirements indicated in Preliminaries, $h_{1}(\tilde{L})$ in heterophilic aggregation and $h_{2}(\tilde{L})$ in homophilic aggregation need to be positive semi-definite, equivalently constraining all eigenvalues of $h_{1}(\tilde{L})$ and $h_{2}(\tilde{L})$ to be non-negative, i.e.,
\begin{equation}
\renewcommand{\arraystretch}{1.25}
\left\{
\begin{array}{l}
h_{1}(\lambda) = \exp(-t (p-2+\lambda))-\alpha_{1} \geq 0 , \forall \lambda \in [0, 2) \\
h_{2}(\lambda) = p-2+\lambda \ge 0 , \forall \lambda \in [0, 2)
\end{array}
\right.
\end{equation}
By solving the inequality above, we obtain the self-loop information measurement $p \in [2,-\frac{1}{t} \ln \alpha_{1})$. For homophilic aggregation, $h_{2}(\tilde{L}) = \tilde{L}$ corresponds to a low-pass filter of $g_{2}(\tilde{L}) = (I + \alpha_{2} \tilde{L})^{-1}$. To avoid matrix inversion, we use the first-order approximation as in the GCN \cite{GCN}. Generally, $k$-th order graph filtering can be written as $g_{2}(\tilde{L}) = (I - \alpha_{2} \tilde{L})^{k}$, where $k$ is a non-negative integer and is related to the degree of smoothing. For simplicity, we set $\alpha_{2} = 1$ in our model. 
We have
\begin{equation}
    Z^*=g_{k,t}(\tilde{L})X=(I-\tilde{L})^{k}e^{t\tilde{L}}X=U \cdot (I-\tilde{\lambda})^{k}e^{t\tilde{\lambda}} \cdot U^{\top}X.
\end{equation}

In fact, BernNet \cite{BernNet} and JacobiConv \cite{JacobiConv} can also be viewed as a combination of local low-pass and local high-pass under two-fold filtering mechanism, i.e., $g(L)=\sum_{a,b}^{K}C_{a b}(2-L)^{a}L^{b}$, where $C_{a b}$ is learnable coefficients with specific initialization. They can be decoupled into  homophilic aggregation  $(2I-L)^{a}$ and  heterophilic aggregation $L^{b}$. Their heterophilic aggregation part is only localized, which may result in the neglect of global homophilic information \cite{Global-hete,fang2022structure,xie2023contrastive}. By contrast, the heterophilic aggregation of PC-Conv is improved to $e^{t \tilde{L}}$, which utilizes the global information in a simple way while also strengthening the importance of even-order neighbors for heterophilic node \cite{EvenNet}.

In view of the exponential form of our spectral graph convolution, we use the Possion-Charlier polynomial \cite{Poisson} to make an exact numerical approximation.
\begin{theorem}
\cite{Poisson-Charlier} The Poisson-Charlier polynomial is the coefficient of the power series expansion of the function $G(\gamma,t, \tilde{\lambda})=(1-\tilde{\lambda})^{\gamma} e^{t \tilde{\lambda}}$ by $\tilde{\lambda}$, i.e.,
\begin{equation}\label{equiveq}
G(\gamma,t, \tilde{\lambda})=\sum_{n=0}^{\infty} \frac{(-\tilde{\lambda})^{n}}{n !} C_{n}(\gamma, t) \quad \gamma, t \in R,\hspace{.1cm} t>0,
\end{equation}
where $C_{n}(\gamma, t)=\sum_{k=0}^{n} C_{n}^{k}(-t)^{k} \gamma \cdots(\gamma-n+k+1)$ is the Poisson-Charlier polynomial and it has the following recurrence relations: 
\begin{equation}
\begin{aligned}
&C_{0}(\gamma, t)=1, C_{1}(\gamma, t)=\gamma- t, ...,  C_{n}(\gamma, t)=(\gamma-n-t+1)\\& C_{n-1}(\gamma, t)- (n-1) t C_{n-2}(\gamma, t), n \geq 2 
\end{aligned}
\label{recursion}
\end{equation}
\end{theorem}

\begin{corollary}
Let $P_{\gamma,t}(\tilde{\lambda})=\sum_{n=0}^{N} \frac{(-\tilde{\lambda})^{n}}{n !} C_{n}(\gamma, t)$ denote the Poisson-Charlier approximation of $G(\gamma,t, \tilde{\lambda})$. We have  $P_{\gamma,t}(\tilde{\lambda}) \rightarrow G(\gamma,t, \tilde{\lambda})$ as $N \rightarrow \infty$. 
\end{corollary}

Using the above approximation, we can obtain our PC-filter as :

\begin{equation}
\begin{aligned}
&g_{k,t}(\tilde{\lambda})=\sum_{n=0}^{N} \frac{(-\tilde{\lambda})^{n}}{n !} C_{n}(k, t), k\in Z_{+},\\
&g_{k,t}(\tilde{\lambda}) \rightarrow (I-\tilde{\lambda})^{k}e^{t\tilde{\lambda}}, N \rightarrow \infty.
\label{pcfilter}
\end{aligned}
\end{equation}
Using a single $k$ for all nodes means that the degree of local information aggregation is fixed, which would lead to an under-smoothing problem \cite{zhang2021node,pan2021multi}. To make full use of node features at multiple orders, we introduce a learnable parameter $\theta_{k}$ to aggregate node information from $k$-order neighbors, where each term can be considered as a single layer filter. 

In addition, for graphs with a high degree of heterophily, topology structure-based aggregation may not learn a good representation. Motivated by \cite{GCN2}, we use identity mapping to incorporate raw node features and ensure that the representation can at least achieve the same performance as the original data.
Finally, we formulate our Possion-Charlier convolution (PC-Conv) as:

\begin{equation}
\begin{aligned} 
Z^* &=g_{t}(\tilde{L})X=U(\theta_{0}+\sum_{k=1}^{K} \theta_{k} \cdot g_{k,t}(\tilde{\lambda}))U^{\top} X\\
    &=\theta_{0} \cdot X +\sum_{k=1}^{K} \theta_{k} \cdot \sum_{n=0}^{N} C_{n}\left(k, t\right) \frac{(-\tilde{L})^{n}}{n !}X
\end{aligned}
\label{pcconv}
\end{equation}
where $C_{n}\left(k, t\right)$ is computed through Eq. (\ref{recursion}). 


\begin{figure}[t]
    \centering
    \subfigure[Comb]{
    \includegraphics[width=39mm]{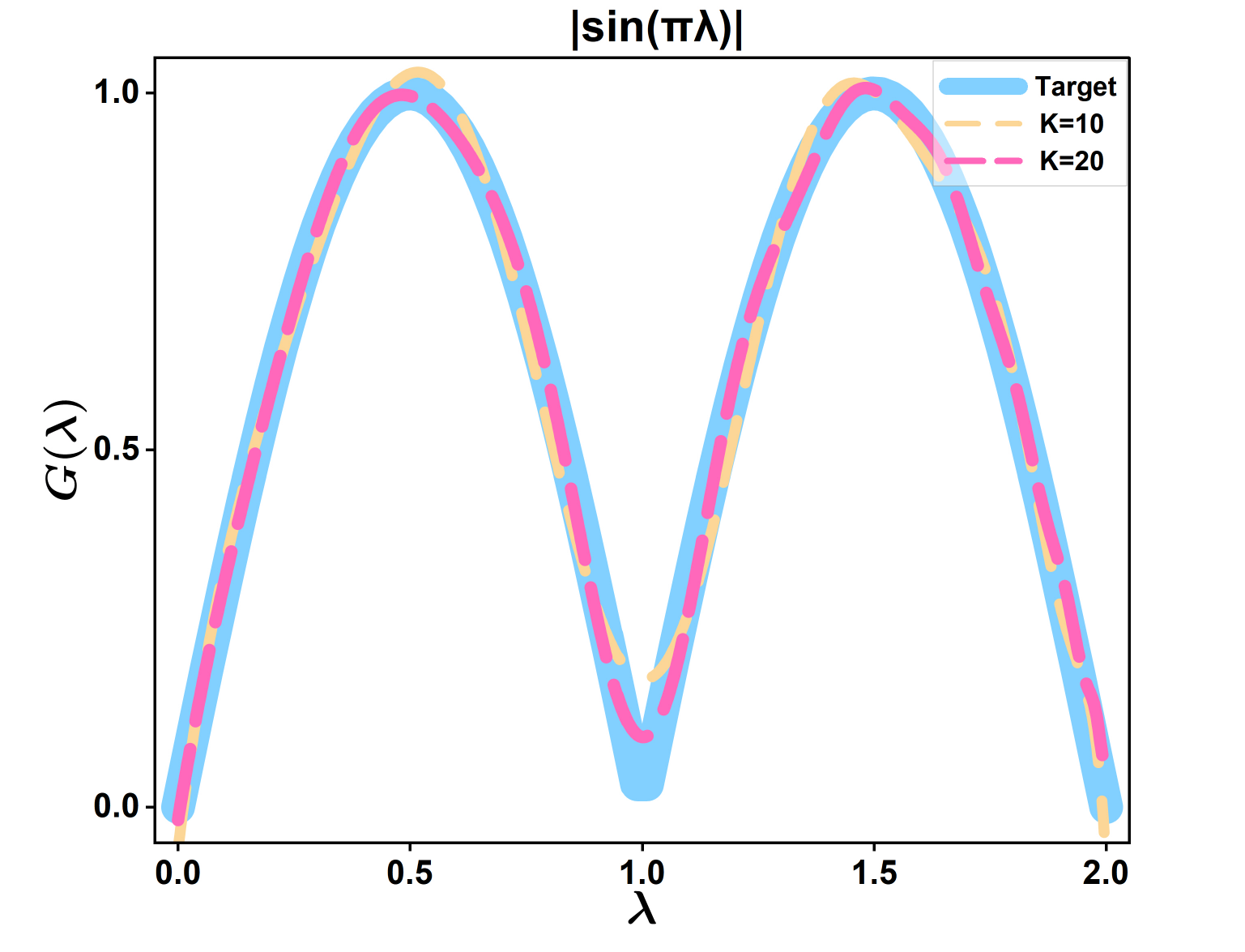}
    }
    \subfigure[Low-band-pass]{
    \includegraphics[width=39mm]{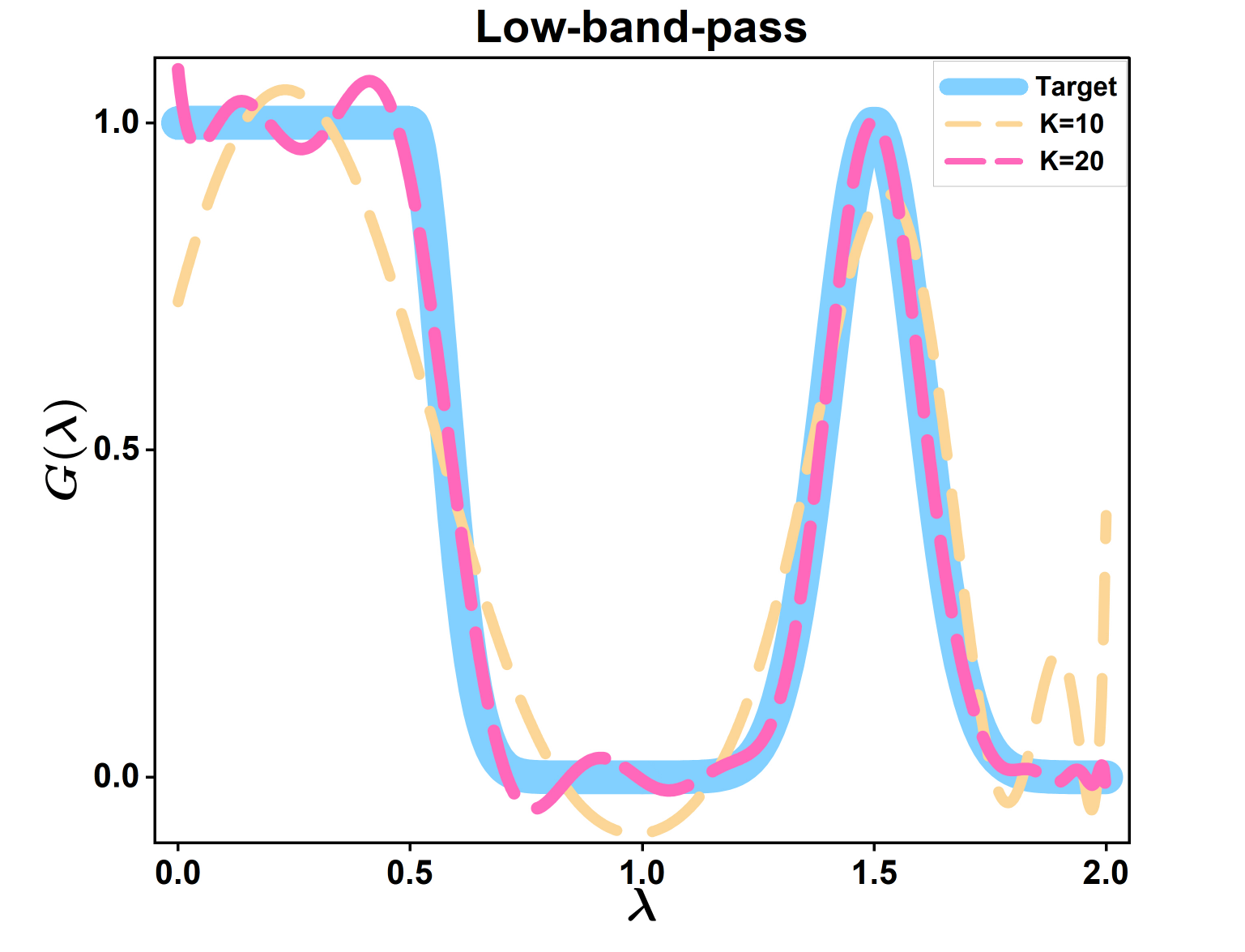}
    }
    \vspace{-2.mm}
     \caption{Illustrations of two complex filters and their approximations learned by PC-Conv.}
     \label{fig:filters}
 \end{figure}

\subsection{Spectral Analysis}
Assuming there exists a perfect filter function $G(\lambda)$ that can best match the downstream task, the objective of filter learning is to approximate $G(\lambda)$ as accurately as possible. 

\begin{theorem}\label{expressive}
When $N \geq K$, $t\not\in\{1,2,\cdots,K\}$, any $N$ order differentiable real value filter function can be approximated by $N$-order PC-Conv, with an error of order $N+1$.
(Proof in Appendix)
\end{theorem}
\begin{corollary}\label{coro}
PC-Conv can learn arbitrary polynomial filters with order up to $N$, where $N \geq K$, $t\not\in\{1,2,\cdots,K\}$.
\end{corollary}

Therefore, PC-Conv belongs to \textit{Polynomial-Filter-Most-Expressive} (PFME) Convolution \cite{JacobiConv}, whose learning ability is positively correlated with $K$. Since the eigenvalues of graph Laplacian matrix are discrete and can be interpolated precisely by polynomials, PC-Conv can be further classified as \textit{Filter-Most-Expressive} (FME) Convolution, which can express arbitrary real-valued filter functions.

Accordingly, PC-Conv can learn more complex filters than low-pass and high-pass filters, such as comb and band-pass filters. In Fig.~\ref{fig:filters}, we plot two difficult filters that are learned by PC-Conv in an end-to-end fashion. The low-band-pass filter is $G(\lambda)=I_{[0,0.5]}(\lambda)+\exp{(-100(\lambda-0. 5)^2)}I_{(0.5,1)}(\lambda ) + \exp{(-50(\lambda-1.5)^2)}I_{[1,2]}(\lambda)$, where $I_{\Omega}(\lambda)=1$ when $\lambda\in\Omega$, otherwise $I_{\Omega}(\lambda)=0$ \cite{BernNet}. We can observe that PC-Conv learns well and the approximation accuracy increases with $K$, which is in line with the above theory. 

\begin{table}[t]
\centering
\caption{The computational complexity of relevant polynomial-based methods. }\label{tab::Computational Complexity}
\begin{center}
\resizebox{0.4\textwidth}{!}{
\begin{tabular}{lc}
\toprule
Model      & Computational Complexity 
\\
\midrule
MLP      &  $ O(m F d + m d C) $  \\
GPRGNN &  $ O (K |E| C) + O(m K C)$ \\
BernNet     &  $O(K^{2} |E| C) + O(m K C) + O(K)$\\
JacobiConv  &   $O(K |E| C) + O(m K C) + O(K)$\\
ChebNetII   &  $O(K |E| C) + O(m K C) + O(K)$  \\
PC-Conv       &   $O(N |E| C) + O(m N K C) + O(K)$   \\
\bottomrule
\end{tabular}}
\end{center}
\vskip -0.15in
\end{table}
\subsection{PCNet for Node Classification}
Based on PC-Conv, we propose PCNet architecture for the node classification task, which contains three components: generalized normalization of the graph Laplacian matrix, the computation of PC-Conv, and  the classification module. 

\cite{Corr2} analyzes the overcorrelation issue in spectral GNNs. 
\cite{Corr} points out that the fixed range $[-1,1]$ of standard Laplacian normalization is not a good choice due to the diminishing problem.
Generalized normalization can alleviate this problem by shrinking the spectrum \cite{Large-pre}. Thus, we introduce the hyperparameter $ \eta$ into PCNet. 
\begin{equation}
    \tilde{A}=(D+ I)^{-\eta}(A+ I)(D+ I)^{-\eta}
\end{equation}
Then, our Laplacian matrix $\tilde{L}$ becomes:
\begin{equation}
    \tilde{L}=(p-1)I-(D+ I)^{-\eta}(A+ I)(D+ I)^{-\eta}
\end{equation}

Then, we calculate the Poisson-Charlie polynomial according to Eq. (\ref{recursion}). Different from existing works, PC polynomial serves as coefficients in our convolution. Therefore, the time complexity of this process is $ O (N)$. The construction of PC-Conv can be seen as two processes. The first step is to calculate the generalized Laplacian matrix for message transmission, i.e., $\tilde{L}^{i}\Theta(X),i\in{{0,\ldots,N}}$, where $\Theta(X)\in R^{m\times C} $ is the transformed feature matrix for $C$ classes, usually an MLP or linear layer. The time complexity is $ O(N|E|C) $. The second step is to combine it with PC coefficients according to Eq. (\ref{pcfilter}),
whose time complexity is $O(mNKC)$. 
 The total computational complexity of our PC-Conv is $O(N|E|C)+O(mNKC)+O(N)$.
The first term dominates the others in practice. We list the computational complexity of relevant methods in Table \ref{tab::Computational Complexity}. Compared to $O(K|E|C)$ in other methods with $K = 10$, PC-Conv is more efficient since  $N$ is often smaller than $K$.
Therefore, our method doesn't incur extra computation burden w.r.t. other SOTA methods.
The node classification with PCNet is:
\begin{equation}
\begin{aligned}
&Z^*=softmax\left((\theta_{0}+\sum_{k=1}^{K} \theta_{k} \cdot g_{k,t}(\tilde{\lambda})) \Theta(X)\right),\\
&\mathcal{L}=-\sum_{l \in \mathcal{Y}_{L}} \sum_{c=1}^{C} Y_{l c} \ln Z_{l c}^*,
\end{aligned}
\end{equation}
where  $\mathcal{Y}_{L}$ is the labeled node indices and the cross-entropy loss over label matrix $Y$ is applied to train the GCN network. The coefficient $\theta_{k}$ is updated by the backward propagation of the loss.

\begin{table*}[t]

\centering
\caption{The results of semi-supervised node classification: Mean accuracy (\%) ± 95\% confidence interval. We highlight the best score in \textcolor{red}{red} and the runner-up score with \textcolor{blue}{blue}.   }
\label{acc_res}
\resizebox{1.0\textwidth}{!}{
\begin{tabular}{l c c c c c c c c c c}
\toprule
Dataset & MLP & GCN & ChebNet & ARMA& APPNP & GPRGNN & BernNet & ChebNetII & PCNet \\ 
\midrule
Cora  

& 57.17$_{\pm\text{1.34}}$ &79.19$_{\pm\text{1.37}}$  &78.08$_{\pm\text{0.86}}$ &79.14$_{\pm\text{1.07}}$ &82.39$_{\pm\text{0.68}}$ &82.37$_{\pm\text{0.91}}$ &82.17$_{\pm\text{0.86}}$
&{\textcolor{blue}{82.42$_{\pm\text{0.64}}$}}
&{\textcolor{red}{82.81$_{\pm\text{0.50}}$}}
\\

CiteSeer 
&56.75$_{\pm\text{1.55}}$ &69.71$_{\pm\text{1.32}}$  &67.87$_{\pm\text{1.49}}$ &69.35$_{\pm\text{1.44}}$ &69.79$_{\pm\text{0.92}}$ &69.22$_{\pm\text{1.27}}$ &69.44$_{\pm\text{0.97}}$
&{\textcolor{blue}{69.89$_{\pm\text{1.21}}$}}
&{\textcolor{red}{69.92$_{\pm\text{0.70}}$}}
\\
PubMed   
&70.52$_{\pm\text{0.27}}$ &78.81$_{\pm\text{0.24}}$ &73.96$_{\pm\text{0.31}}$ &78.31$_{\pm\text{0.22}}$ 
&{\textcolor{blue}{79.97$_{\pm\text{0.28}}$}}
&79.28$_{\pm\text{0.33}}$  &79.48$_{\pm\text{0.41}}$ &79.51$_{\pm\text{1.03}}$
&{\textcolor{red}{80.01$_{\pm\text{0.88}}$}}
\\
Texas
&32.42$_{\pm\text{9.91}}$ &34.68$_{\pm\text{9.07}}$   &36.35$_{\pm\text{8.90}}$   &39.65$_{\pm\text{8.09}}$  &34.79$_{\pm\text{10.11}}$ &33.98$_{\pm\text{11.90}}$ &43.01$_{\pm\text{7.45}}$ 
&{\textcolor{blue}{46.58$_{\pm\text{7.68}}$}}
&{\textcolor{red}{64.56$_{\pm\text{1.84}}$}}
\\

Actor
 &29.75$_{\pm\text{0.95}}$  &22.74$_{\pm\text{2.37}}$  &26.58$_{\pm\text{1.92}}$ &27.02$_{\pm\text{2.31}}$  &29.74$_{\pm\text{1.04}}$  &28.58$_{\pm\text{1.01}}$ 
&29.87$_{\pm\text{0.78}}$  
 &{\textcolor{blue}{30.18$_{\pm\text{0.81}}$}}
 &{\textcolor{red}{33.56$_{\pm\text{0.40}}$}}
\\


Cornell
&36.53$_{\pm\text{7.92}}$ &32.36$_{\pm\text{8.55}}$ &28.78$_{\pm\text{4.85}}$  &28.90$_{\pm\text{10.07}}$  &34.85$_{\pm\text{9.71}}$  &38.95$_{\pm\text{12.36}}$ &39.42$_{\pm\text{9.59}}$
&{\textcolor{blue}{42.19$_{\pm\text{11.61}}$}}
&{\textcolor{red}{52.08$_{\pm\text{4.45}}$}}
\\

\bottomrule
\end{tabular}}
\end{table*}
\section{Experiment}
\begin{table*}[t]
\centering
\caption{Mean accuracy (\%) of node classification with polynomial-based methods.}
\resizebox{0.7\textwidth}{!}{
\begin{tabular}{@{}llll lll@{}}
\toprule
  Method & Cora & Citeseer & Pubmed &Texas  &Actor &Cornell   \\ \midrule
MLP 
&76.89$_{\pm\text{0.97}}$
&76.52$_{\pm\text{0.89}}$
&86.14$_{\pm\text{0.25}}$
&86.81$_{\pm\text{2.24}}$

&40.18$_{\pm\text{0.55}}$

 &84.15$_{\pm\text{3.05}}$

\\

GCN        
&87.18$_{\pm\text{1.12}}$      &79.85$_{\pm\text{0.78}}$          &86.79$_{\pm\text{0.31}}$
&76.97$_{\pm\text{3.97}}$ 
&33.26$_{\pm\text{1.15}}$
&65.78$_{\pm\text{4.16}}$

\\

ChebNet    
&87.32$_{\pm\text{0.92}}$      &79.33$_{\pm\text{0.57}}$          &87.82$_{\pm\text{0.24}}$
&86.28$_{\pm\text{2.62}}$

&37.42$_{\pm\text{0.58}}$
&83.91$_{\pm\text{2.17}}$

 \\

ARMA      
&87.13$_{\pm\text{0.80}}$      &80.04$_{\pm\text{0.55}}$          &86.93$_{\pm\text{0.24}}$
&83.97$_{\pm\text{3.77}}$
 
&37.67$_{\pm\text{0.54}}$
&85.62$_{\pm\text{2.13}}$

 \\

APPNP      
&88.16$_{\pm\text{0.74}}$      &80.47$_{\pm\text{0.73}}$
&88.13$_{\pm\text{0.33}}$  
&90.64$_{\pm\text{1.70}}$ 
&39.76$_{\pm\text{0.49}}$
&91.52$_{\pm\text{1.81}}$

\\

GCNII        
&88.46$_{\pm\text{0.82}}$      &79.97$_{\pm\text{0.65}}$   
&\textcolor{blue}{89.94$_{\pm\text{0.31}}$}
&80.46$_{\pm\text{5.91}}$
&36.89$_{\pm\text{0.95}}$
&84.26$_{\pm\text{2.13}}$

\\

TWIRLS 
&88.57$_{\pm\text{0.91}}$
&80.07$_{\pm\text{0.94}}$          &88.87$_{\pm\text{0.43}}$  
&91.31$_{\pm\text{3.36}}$
&38.13$_{\pm\text{0.81}}$
&89.83$_{\pm\text{2.29}}$

 \\

EGNN
&87.47$_{\pm\text{1.33}}$
&80.51$_{\pm\text{0.93}}$         &88.74$_{\pm\text{0.46}}$
&81.34$_{\pm\text{1.56}}$
&35.16$_{\pm\text{0.64}}$
&82.09$_{\pm\text{1.16}}$

\\

PDE-GCN

&88.62$_{\pm\text{1.03}}$
&79.98$_{\pm\text{0.97}}$      &89.92$_{\pm\text{0.38}}$
&93.24$_{\pm\text{2.03}}$

&39.76$_{\pm\text{0.74}}$
&89.73$_{\pm\text{1.35}}$

\\

GPRGNN     
&88.54$_{\pm\text{0.67}}$      &80.13$_{\pm\text{0.84}}$          &88.46$_{\pm\text{0.31}}$
&92.91$_{\pm\text{1.32}}$
&39.91$_{\pm\text{0.62}}$ 
&91.57$_{\pm\text{1.96}}$
 \\

BernNet     
&88.51$_{\pm\text{0.92}}$      &80.08$_{\pm\text{0.75}}$          &88.51$_{\pm\text{0.39}}$
&92.62$_{\pm\text{1.37}}$
&41.71$_{\pm\text{1.12}}$
&92.13$_{\pm\text{1.64}}$
\\

EvenNet
&87.25$_{\pm\text{1.42}}$
&78.65$_{\pm\text{0.96}}$
&89.52$_{\pm\text{0.31}}$
&\textcolor{blue}{93.77$_{\pm\text{1.73}}$}
&40.48$_{\pm\text{0.62}}$
&92.13$_{\pm\text{1.72}}$
\\

ChebNetII 
&88.71$_{\pm\text{0.93}}$      
&80.53$_{\pm\text{0.79}}$      
&88.93$_{\pm\text{0.29}}$
&93.28$_{\pm\text{1.47}}$
 &{\textcolor{blue}{41.75$_{\pm\text{1.07}}$}}

&92.30$_{\pm\text{1.48}}$

\\ 
JacobiConv   
&\textcolor{blue}{88.98$_{\pm\text{0.46}}$}
&\textcolor{blue}{80.78$_{\pm\text{0.79}}$}
&89.62$_{\pm\text{0.41}}$

&93.44$_{\pm\text{2.13}}$
&41.17$_{\pm\text{0.64}}$

&\textcolor{blue}{92.95$_{\pm\text{2.46}}$}

\\

PCNet 

  &{\textcolor{red}{90.02$_{\pm\text{0.62}}$}}
 &{\textcolor{red}{81.76$_{\pm\text{0.78}}$}}
 &{\textcolor{red}{91.30$_{\pm\text{0.38}}$}}
  &{\textcolor{red}{95.57$_{\pm\text{1.31}}$}}
  &{\textcolor{red}{42.72$_{\pm\text{0.91}}$}}
 &{\textcolor{red}{93.83$_{\pm\text{1.91}}$}}

\\
\bottomrule
\end{tabular}
\label{tb:full}}
\end{table*}
{\bf Dataset}~~

For the sake of fairness, we select benchmark datasets that are commonly used in related work. 
For the homophilic graph, we select three classic citation graphs: Cora, CiteSeer, PubMed \cite{Cora}. For the heterophilic graph, we select four datasets including the Webpage graphs Texas, Cornell, and Wisconsin from WebKB3 \footnote{http://www.cs.cmu.edu/afs/cs.cmu.edu/project/theo-11/www/wwkb}, and the Actor co-occurrence graph \cite{Gemo}. Besides, we also select Penn94 from Facebook 100 \footnote{https://archive.org/details/oxford-2005-facebook-matrix} as a representative dataset of the social network \cite{LINKX}. 

\noindent{\bf Train/validation/test split}~~

For semi-supervised node classification, we split the dataset with random seeds according to \cite{Chebyshev2}. In detail, we select 20 nodes of each class on three homophilic datasets (Cora, Citeser, and Pubmed) for training, 500 nodes for validation, and 1000 nodes for testing. For the three heterophilic datasets (Texas, Actor, Cornell), we apply the sparse splitting where the training/validation/test
sets account for 2.5\%/2.5\%/95\%.

For node classification, there are two common split ways: random seed splits and fixed seed splits. The former, proposed by \cite{GPRGNN}, uses random seeds for random split and is widely used in spectral GNNs, PDE GNNs, traditional methods, etc. The latter uses the fixed split given in \cite{Global-hete}, which is commonly used in heterophilic GNNs.  To avoid unfair comparison, we evaluate the performance of PCNet under these two different settings. 
For the node classification with polynomial-based methods, we adopt 10 random seeds as \cite{GPRGNN,JacobiConv} to randomly split the dataset into train/validation/test sets with a ratio of 60\%/20\%/20\%. For node classification with heterophilic GNNs, we use the given split in  \cite{Global-hete} to randomly split the dataset into training/validation/test sets at a ratio of  48\%/32\%/20\%.

\begin{table*}[t]
\centering
\caption{Mean accuracy (\%) of node classification by the heterophilic methods. OOM refers to the out-of-memory error.}

\label{tab:result_heter}
\resizebox{1.0\textwidth}{!}{
\begin{tabular}{@{}lllll lllll@{}}
\toprule
Dataset & Cora & Citeseer & Pubmed & Texas & Actor & Cornell & Wisconsin &Penn94 &Avg.Rank\\ \midrule
     MLP     
     & 75.69$_{\pm\text{2.00}} $   
     & 74.02$_{\pm\text{1.90}} $  
     & 87.16$_{\pm\text{0.37}} $                
     & 80.81$_{\pm\text{4.75}} $       
     & 36.53$_{\pm\text{0.70}} $  
     & 81.89$_{\pm\text{6.40}} $
     & 85.29$_{\pm\text{3.31}} $
     & 73.61$_{\pm\text{0.40}} $
     &9.19

 \\
     GCN         
     & 86.98$_{\pm\text{1.27}} $ 
     & 76.50$_{\pm\text{1.36}} $    
     & 88.42$_{\pm\text{0.50}} $            
     & 55.14$_{\pm\text{5.16}} $          
     & 27.32$_{\pm\text{1.10}} $ 
     & 60.54$_{\pm\text{5.30}} $ 
     & 51.76$_{\pm\text{3.06}} $
     & 82.47$_{\pm\text{0.27}} $
     &10.75
\\
     GAT       
     & 87.30$_{\pm\text{1.10}} $ 
     & 76.55$_{\pm\text{1.23}} $ 
     & 86.33$_{\pm\text{0.48}} $             
     & 52.16$_{\pm\text{6.63}} $          
     & 27.44$_{\pm\text{0.89}} $ 
     & 61.89$_{\pm\text{5.05}} $  
     & 49.41$_{\pm\text{4.09}} $
     & 81.53$_{\pm\text{0.55}} $
     &11.13

 \\    
      MixHop      
      & 87.61$_{\pm\text{0.85}} $ 
      & 76.26$_{\pm\text{1.33}} $ 
      & 85.31$_{\pm\text{0.61}} $          
      & 77.84$_{\pm\text{7.73}} $         
      & 32.22$_{\pm\text{2.34}} $ 
      & 73.51$_{\pm\text{6.34}} $ 
      & 75.88$_{\pm\text{4.90}} $
      & 83.47$_{\pm\text{0.71}} $
      &9.75


  

\\
   
     GCNII     
     & \textcolor{blue}{88.37$ _{\pm\text{1.25}}$}  
     & \textcolor{blue}{77.33$ _{\pm\text{1.48}}$} 
     & \textcolor{red}{90.15$ _{\pm\text{0.43}}$ }       
     & 77.57 $_{\pm\text{3.83}} $          
     & 37.44$ _{\pm\text{1.30}} $  
     & 77.86$ _{\pm\text{3.79}} $
     & 80.39$_{\pm\text{3.40}} $
     & 82.92$_{\pm\text{0.59}} $
     &5.25

 \\
    H$_2$GCN     
    & 87.87$_{\pm\text{1.20}} $ 
    & 77.11$_{\pm\text{1.57}} $  
    & 89.49$_{\pm\text{0.38}} $           
    & 84.86$_{\pm\text{7.23}} $          
    & 35.70$_{\pm\text{1.00}} $  
    & 82.70$_{\pm\text{5.28}} $ 
    & 87.65$_{\pm\text{4.98}} $
    & 81.31$_{\pm\text{0.60}} $
    &6.19

 \\ 
    WRGAT     
    & 88.20$_{\pm\text{2.26}} $ 
    & 76.81$_{\pm\text{1.89}}  $  
    & 88.52$_{\pm\text{0.92}} $          
    & 83.62$_{\pm\text{5.50}}  $         
    & 36.53$_{\pm\text{0.77}}  $
    & 81.62$_{\pm\text{3.90}} $  
    & 86.98$_{\pm\text{3.78}} $
    & 74.32$_{\pm\text{0.53}} $
    &6.69

 \\ 
    GPRGNN     
    & 87.95$_{\pm\text{1.18}} $ 
    & 77.13$_{\pm\text{1.67}} $ 
    & 87.54$_{\pm\text{0.38}} $          
    & 78.38$_{\pm\text{4.36}} $        
    & 34.63$_{\pm\text{1.22}} $  
    & 80.27$_{\pm\text{8.11}} $ 
    & 82.94$_{\pm\text{4.21}} $
    & 81.38$_{\pm\text{0.16}} $
    &8.06

  \\ 
    GGCN      
    & 87.95$_{\pm\text{1.05}} $ 
    & 77.14$_{\pm\text{1.45}} $
    & 89.15$_{\pm\text{0.37}} $            
    & 84.86$_{\pm\text{4.55}} $        
    & 37.54$_{\pm\text{1.56}} $ 
    & \textcolor{blue}{85.68$_{\pm\text{6.63}} $}
    & 86.86$_{\pm\text{3.22}} $
    & OOM
    &5.50

  \\ 
       ACM-GCN          
        & 87.91$_{\pm\text{0.95}} $
        & 77.32$_{\pm\text{1.70}}  $ 
        & \textcolor{blue}{90.00$_{\pm\text{0.52}} $}
        & \textcolor{blue}{87.84 $_{\pm\text{4.40}}$ } 
        & 36.28$_{\pm\text{1.09}}  $   
        & 85.14$_{\pm\text{6.07}}  $ 
        & \textcolor{blue}{88.43$_{\pm\text{3.22}} $}
        & 82.52$_{\pm\text{0.96}} $
        &4.00

   \\  
       LINKX      
       & 84.64$_{\pm\text{1.13}} $ 
       & 73.19$_{\pm\text{0.99}}  $ 
       & 87.86$_{\pm\text{0.77}} $          
       & 74.60$_{\pm\text{8.37}} $        

       & 36.10$_{\pm\text{1.55}} $ 

       & 77.84$_{\pm\text{5.81}}$ 
       & 75.49$_{\pm\text{5.72}}$ 
       & 84.71$_{\pm\text{0.52}} $
       &9.63

  \\
      GloGNN++   
      & 88.33$_{\pm\text{1.09}} $ 
      & 77.22$_{\pm\text{1.78}} $ 
      & 89.24$_{\pm\text{0.39}} $ 
      & 84.05$_{\pm\text{4.90}}$       
      & \textcolor{blue}{37.70$_{\pm\text{1.40}}$ }

      & \textcolor{red}{85.95$_{\pm\text{5.10}}$ } 
       & 88.04$_{\pm\text{3.22}}$ 
       & \textcolor{red}{85.74$_{\pm\text{0.42}}$ }
       &\textcolor{blue}{3.00 }

   \\  
       PCNet    
        & \textcolor{red}{88.41$_{\pm\text{0.66}}$ } 
        & \textcolor{red}{77.50$_{\pm\text{1.06}}$  }
        & 89.51$_{\pm\text{0.28}}$  
        & \textcolor{red}{88.11$_{\pm\text{2.17}} $}  

        & \textcolor{red}{37.80$_{\pm\text{0.64}}$ }  

        & 82.16$_{\pm\text{2.70}} $
        & \textcolor{red}{88.63$_{\pm\text{2.75}} $}
        & \textcolor{blue}{84.75$_{\pm\text{0.51}} $}
        &\textcolor{red}{1.88}

\\
 
\bottomrule

\end{tabular}}
\end{table*}

\begin{table*}[t]
\centering
\caption{Results of ablation study on PC-Conv with MLP and single linear layer.}\label{tab::abl1}
\begin{center}
\resizebox{1.0\textwidth}{!}{
\begin{tabular}{ l ccccc | ccccc }

\toprule
Datasets & Monomial & Chebyshev & Bernstein & Jacobi   &PCNet$_{0.5}$-Lin       &GPRGNN &ChebNet &BernNet  &Jacobi-MLP &PCNet$_{0.5}$\\
\midrule

Cora 
&88.80$_{\pm\text{0.67}}$ 
&88.49$_{\pm\text{0.82}}$ 
&86.50$_{\pm\text{1.26}}$ 
&88.98$_{\pm\text{0.72}}$ 
&\textcolor{red}{90.04$_{\pm\text{ 0.71}}$}
&88.54$_{\pm\text{0.67}}$ 
&87.32$_{\pm\text{0.92}}$   
&88.51$_{\pm\text{0.92}}$ 
&88.67$_{\pm\text{0.69}}$
&\textcolor{red}{90.00$_{\pm\text{0.49}}$}
\\
Citeseer
&80.68$_{\pm\text{0.86}}$ 
&80.53$_{\pm\text{0.81}}$ 
&80.61$_{\pm\text{0.85}}$ 
&80.61$_{\pm\text{0.72}}$ 
&\textcolor{red}{81.17$_{\pm\text{0.66}}$}
&80.13$_{\pm\text{0.84}}$
&79.33$_{\pm\text{0.57}}$
&80.08$_{\pm\text{0.75}}$ 
&80.25$_{\pm\text{0.60}}$ 
&\textcolor{red}{82.10$_{\pm\text{0.72}}$}

 \\
Pubmed 
&89.54$_{\pm\text{0.36}}$ 
&89.52$_{\pm\text{0.46}}$ 
&88.42$_{\pm\text{0.32}}$ 
&89.70$_{\pm\text{0.34}}$ 
&\textcolor{red}{90.97$_{\pm\text{0.27}}$}
&88.46$_{\pm\text{0.31}}$
&87.82$_{\pm\text{0.24}}$
&88.51$_{\pm\text{0.92}}$ 
&87.73$_{\pm\text{2.13}}$ 
&\textcolor{red}{91.58$_{\pm\text{0.95}}$}
\\
Texas 
&91.64$_{\pm\text{2.46}}$ 
&88.36$_{\pm\text{3.93}}$ 
&89.34$_{\pm\text{2.46}}$ 
&92.79$_{\pm\text{1.97}}$ 
&\textcolor{red}{95.25$_{\pm\text{1.15}}$}
&92.91$_{\pm\text{1.32}}$
&87.82$_{\pm\text{0.24}}$
&92.62$_{\pm\text{1.37}}$
&89.34$_{\pm\text{3.12}}$
&\textcolor{red}{95.25$_{\pm\text{0.82}}$}


\\
Actor 
&40.31$_{\pm\text{0.82}}$ 
&40.61$_{\pm\text{0.64}}$ 
&40.42$_{\pm\text{0.50}}$  
&\textcolor{red}{40.70$_{\pm\text{0.98}}$}
&39.56$_{\pm\text{1.00}}$ 
&39.91$_{\pm\text{0.62}}$
&37.42$_{\pm\text{0.58}}$
&41.71$_{\pm\text{1.12}}$
&37.56$_{\pm\text{0.88}}$ 
&\textcolor{red}{41.79$_{\pm\text{0.70}}$}
\\

Cornell 
&91.31$_{\pm\text{2.13}}$ 
&88.03$_{\pm\text{3.28}}$ 
&92.46$_{\pm\text{2.63}}$ 
&92.30$_{\pm\text{2.79}}$ 
&\textcolor{red}{92.62$_{\pm\text{1.97}}$}
&91.57$_{\pm\text{1.96}}$
&83.91$_{\pm\text{2.17}}$
&92.13$_{\pm\text{1.64}}$
&87.54$_{\pm\text{3.11}}$
&\textcolor{red}{92.32$_{\pm\text{1.48}}$}
\\
\bottomrule
\end{tabular}}

\end{center}
\end{table*}

\subsection{Semi-supervised Node Classification}
\textbf{Settings:} We compare the performance of PCNet with eight polynomial approximation methods of spectral filtering, including MLP, GCN \cite{GCN}, APPNP \cite{APPNP}, ChebNet \cite{ChebyNet}, ARMA 
 \cite{ARMA}, GPRGNN \cite{GPRGNN}, BernNet \cite{BernNet}, and ChebNetII \cite{Chebyshev2}. For PCNet, we use the same settings for all experiments. Specifically, we apply MLP to dimensionally process the features and set the hidden units to 64 for fairness  \cite{GPRGNN,Chebyshev2}. To reduce computation load, we set $K = 4\sim 6$. 

\textbf{Result:} We can observe that PCNet outperforms all other methods in Table \ref{acc_res}.
PCNet even achieves a performance improvement of $10\% \sim20\%$ on two heterophilic datasets: Texas and Cornell. In fact, under sparse split, the performance of many methods fluctuates a lot with the choice of the training set. By contrast, the fluctuations in PCNet are small, which reflects the stability of PCNet. Furthermore, PCNet uses only $4\sim$ 6 PC-filters, achieving competitive performance yet saving computational overhead.

\subsection{Node Classification with Polynomial-based Methods}\label{exper5.2}
\textbf{Settings: }
Besides JacobiConv \cite{JacobiConv} and EvenNet \cite{EvenNet}, we also include four competitive baselines for node classification: GCNII \cite{GCN2}, TWIRLS \cite{TWIRLS}, EGNN \cite{EGNN}, and PDE-GCN \cite{PDEGCN}, which make full use of the topology information from different perspectives including GNNs architecture, energy function, and PDE GNNs. 

\textbf{Result:} 
Table \ref{tb:full} shows that PCNet achieves the best performance on all datasets and has at least 1\% gain w.r.t. the second best performance.
PCNet's excellent performance on both homophilic and heterophilic graphs is due to our two-fold filtering which captures the co-existence of homophily and heterophily in real datasets. In addition, the heterophilic heat kernel also contributes to this superior performance by aggregating more distant (and even infinite-order) neighbor information compared to BernNet and JacobiConv.

\subsection{Node Classification with Heterophilic GNNs}\label{exper5.3}

\textbf{Settings:} We add some heterophilic methods, including (1) Spatial heterophilic GNNs: H2GCN \cite{H2GCN}, WRGAT \cite{WRGAT},  GloGNN++ \cite{Global-hete}; (2) GNNs with filterbank:  GPRGNN \cite{GPRGNN} and ACM-GCN \cite{ACM-GCN}; (3) MLP-based methods: LINKX \cite{LINKX}; (4) Scalable heterophilic GNNs: GCNII \cite{GCN2} and GGCN \cite{GGCN}.

\textbf{Result:} 
PCNet achieves the top average ranking on all datasets in Table  \ref{tab:result_heter}. In particular, it outperforms recent heterophilic methods GloGNN++ and GGCN on 6 out of 8 datasets. The superior performance is attributed to our two-fold filtering, which not only explores global information but also mines homophily in heterophilic graphs. Thus PCNet generalizes well to both homophilic and heterophilic graphs. We also see the importance of node features, which provide rich information different from heterophilic structure. Thus 
 MLP can outperform GCN and GAT by more than 20\% in Texas and Cornell. This validates the necessity of identity mapping.

\subsection{Ablation Study on PC-Conv}\label{exper5.4}

To examine the contribution of generalized Laplacian normalization, we replace $\eta$ with 0.5, i.e., we use traditional Laplacian. The   classification results on six datasets are 82.33$_{\pm\text{0.83}}$, 69.01$_{\pm\text{1.08}}$, 79.73 $_{\pm\text{0.97}}$, 57.55$_{\pm\text{2.66}}$ , 33.28$_{\pm\text{0.63}}$, 48.03$_{\pm\text{7.69}}$. Though they outperform other methods in Table \ref{acc_res}, they are still lower than that of PCNet. Thus, it is beneficial to introduce spectral scaling.

Furthermore, to test the effect of PC-Conv without relying on any trick (generalized Laplacian normalization),
we compare popular polynomial bases (Monomial \cite{GPRGNN}, Chebyshev \cite{ChebyNet}, Bernstein \cite{BernNet}, Jacobi \cite{JacobiConv}) to demonstrate the superiority of PC-Conv. In order to exclude the effect of dimensional transformations, both MLP and individual linear layers are applied, and the corresponding models are denoted as PCNet$_{0.5}$ and PCNet$_{0.5}$-LIN respectively. The results on node classification are shown in Table \ref{tab::abl1}, from which we can see that PCNet$_{0.5}$ and PCNet$_{0.5}$-LIN can still beat others in most cases. Moreover, the performance of PCNet$_{0.5}$ and PCNet$_{0.5}$-LIN is close to each other in 5 out of 6 datasets, indicating that PC-Conv is robust to the dimensional transformation.

Finally, we remove the identity mapping from PC-Conv, leaving the combinations of PC-filters. The classification performance on those six datasets are 89.92$_{\pm\text{0.95}}$, 81.01$_{\pm\text{0.63}}$, 90.96$_{\pm\text{0.35}}$, 95.25$_{\pm\text{2.30}}$, 42.11$_{\pm\text{0.92}}$, 93.44$_{\pm\text{2.13}}$. We can observe that the accuracy is inferior to that of PCNet in Table~\ref{acc_res}. But compared to other SOTA methods, they are still better, which illustrates the superiority of the PC-filter itself.

\section{Conclusion}
Motivated by the coexistence of homophilic and heterophilic nodes in real graphs, we propose a two-fold filtering mechanism to simultaneously perform homophilic and heterophilic information aggregation. For the first time, we extend the graph heat kernel to heterophilic graphs to exploit the global information. The proposed PC-filter makes full use of data information and PC-Conv performs adaptive graph convolution. The PCNet architecture for node classification achieves state-of-the-art performance on real-world datasets. One limitation is that our knowledge of homophily is limited, thus we just use the simplest low-pass filter, which deserves more attention in the future.


\bibliography{aaai24}
\clearpage

\begin{center}
	{\large{\bf Appendix}}
\end{center}

\appendix


\section{Proof of Property \ref{order-invariant}}
 $Z^*$ is order-invariant if we exchange the heterophilic and homophilic aggregation operations. 
 \begin{proof}
\begin{equation}
\begin{aligned}
\setlength{\arraycolsep}{1.0pt}
 Z^*
 &=[(\alpha_{1} I+ h_{1}(L))(\alpha_{2} I+ h_{2}(L))]^{-1} X\\
  &=(\alpha_{2} I+ h_{2}(L))^{-1}(\alpha_{1} I+ h_{1}(L))^{-1} X\\
 &=U(\alpha_{2} I + h_{2}(\Lambda))^{-1}(\alpha_{1} I + h_{1}(\Lambda))^{-1}U^{\top}X\\
 &=U(\alpha_{1} I + h_{1}(\Lambda))^{-1}(\alpha_{2} I + h_{2}(\Lambda))^{-1}U^{\top}X\\
&=[(\alpha_{2} I+ h_{2}(L))(\alpha_{1} I+ h_{1}(L))]^{-1} X.
\end{aligned}
\end{equation}
\end{proof}

\section{Proof of Theorem \ref{expressive}}
When $N \geq K$, $t\not\in\{1,2,\cdots,K\}$, any $N$-order differentiable real-valued filter function can be approximated by $N$-order PC-Conv, with an error of order $N+1$.
\begin{proof}
Any $N$-order differentiable filter function $G(\lambda)$, $\lambda\in [0,2)$, can be expanded by a $N$-order Taylor expansion, i.e.,
$G(\lambda)=\sum_{n=0}^{N}G^{(n)}(\lambda) \frac{\lambda^{n}}{n!}+O(\lambda^{N+1})$.  The learning process of PC-filter $g_{k,t}(\lambda)$ via $N$-order Taylor expansion of $G(\lambda)$ can be expressed as:
\begin{align*}
g_{t}(\lambda) &= \theta_{0}+\sum_{k=1}^{K} \theta_{k} \cdot g_{k,t}(\lambda) =\theta_{0}+\sum_{k=1}^{K} \theta_{k} \sum_{n=0}^{N}  C_{n}\left(k, t\right) \frac{(-\lambda)^{n}}{n !}\\
&=\theta_{0}+\sum_{n=0}^{N} \sum_{k=1}^{K} \theta_{k} C_{n}\left(k, t\right) \frac{(-\lambda)^{n}}{n !}
= \sum_{n=0}^{N}G^{(n)}(\lambda) \frac{\lambda^{n}}{n!}\\
&\approx  G(\lambda)
\end{align*}
We have
\begin{equation}
\setlength{\arraycolsep}{1.0pt}
\begin{cases}
\sum_{k=1}^{K} \theta_{k} C_{0}\left(k, t\right) = G^{(0)}(\lambda)-\theta_{0}, n=0  \\
\sum_{k=1}^{K} \theta_{k} C_{n}\left(k, t\right) = (-1)^{n}G^{(n)}(\lambda), n=1,...,N
\end{cases}
\end{equation}
i.e.,
\begin{equation}\label{vand}
\setlength{\arraycolsep}{1.0pt}
    \begin{bmatrix} 
    & C_{0}(1,t) & C_{0}(2,t)&\cdots &C_{0}(K,t) \\&C_{1}(1,t) & C_{1}(2,t)  &\cdots &C_{1}(K,t)\\
    & C_{2}(1,t) & C_{2}(2,t) &\cdots &C_{2}(K,t) \\
    &\vdots &\vdots &\vdots&\vdots\\
     &C_{N}(1,t) & C_{N}(2,t) &\cdots &C_{N}(K,t)
    \end{bmatrix}
    \begin{bmatrix} 
    \theta_1\\\theta_2\\\vdots\\\theta_K
    \end{bmatrix}=
    \begin{bmatrix} 
    G^{(0)}(\lambda)-\theta_{0}\\-G^{(1)}(\lambda)\\\vdots\\(-1)^{N}G^{(N)}(\lambda)
    \end{bmatrix}
\end{equation}

Denote the above equation as $A\cdot\theta=b$, $A\in R^{(N+1) \times K}, \theta\in R^{K}, b\in R^{N+1}$. We know that $G(\lambda)$ is arbitrary on different datasets or downstream tasks. Thus, the problem is equivalent to proving that there exists a set of coefficients $\{\theta_{1},\cdots,\theta_{K}\}$ that can make a linear combination of $A=\{A_{(1)},\cdots,A_{(K)}\}$ into arbitrary vector $b$. We only need to show that $A$ is maximal linearly independent, i.e., the individual column vectors $\{A_{(1)},\cdots,A_{(K)}\}$ are linearly independent.

First, we assume that the individual column vectors $\{A_{(1)},\cdots,A_{(K)}\}$ are linearly correlated. Then there exists a set of coefficients that are not all zeros $\{\alpha_{1},\cdots,\alpha_{K}\}$, such that $\alpha_{1}A_{(1)}+\cdots+\alpha_{K}A_{(K)}=0$.
\begin{align*}
\setlength{\arraycolsep}{1.0pt}
    \alpha_{1}C_{0}(1,t)+&\cdots+\alpha_{K}C_{0}(K,t)=0\\ 
    \alpha_{1}C_{1}(1,t)+&\cdots+\alpha_{K}C_{1}(K,t)=0\\
            &\vdots\\
    \alpha_{1}C_{N-1}(1,t)+&\cdots+\alpha_{K}C_{N-1}(K,t)=0 \\       
    \alpha_{1}C_{N}(1,t)+&\cdots+\alpha_{K}C_{N}(K,t)=0
\end{align*}
i.e.,
\begin{equation}\label{expa}
\setlength{\arraycolsep}{1.0pt}
    \begin{bmatrix} & \alpha_{1}C_{0}(1,t) & \alpha_{2}C_{0}(2,t)&\cdots &\alpha_{K}C_{0}(K,t) \\&\alpha_{1}C_{1}(1,t)&  \alpha_{2}C_{1}(2,t)  &\cdots &\alpha_{K}C_{1}(K,t)\\
    & \alpha_{1}C_{2}(1,t) & \alpha_{2}C_{2}(2,t) &\cdots &\alpha_{K}C_{2}(K,t) \\
    &\vdots &\vdots &\vdots\\
     &\alpha_{1}C_{N}(1,t) &\alpha_{2} C_{N}(2,t) &\cdots &\alpha_{K}C_{N}(K,t)
    \end{bmatrix}
    \begin{bmatrix} 
    1\\1\\\vdots\\1
    \end{bmatrix}=
    \begin{bmatrix} 
    0\\0\\0\\\vdots\\0
    \end{bmatrix}
\end{equation}
Denote the above equation as $B\cdot\mathbf{1}=\mathbf{0}, B\in R^{(N+1) \times K}, \mathbf{1}\in R^{K}, \mathbf{0}\in R^{N+1}$.
Second, define $\delta(x) = max\{x,0\}$ and $P_{i}\in \mathbb{R}^{(N+1)\times(N+1)}$.
\begin{equation}
\setlength{\arraycolsep}{1.0pt}
    (P_{i})_{m n}=
\begin{cases}
1\quad  m = n = 1,\cdots,N+1\\
\delta(m-2-i)t \quad m = n-2, n\geq 3 \\ 
\delta(m-2-i)+t \quad m = n-2, n\geq 3 \\ 
0\quad \textit{others}
\end{cases}
\end{equation}
Left multiplying both sides of Eq. (\ref{expa}) simultaneously by $P_{N-1} , \cdots , P_{0}$ and employing the relation
$C_{n+1}(\alpha, t)=(\alpha-n-t) C_{n}(\alpha, t)-n t C_{n-1}(\alpha, t), n \geq 1$, we have
\begin{equation}
\setlength{\arraycolsep}{1.0pt}
    \begin{bmatrix} & \alpha_{1}C_{0}(1,t) & \alpha_{2}C_{0}(2,t)&\cdots &\alpha_{K}C_{0}(K,t) \\&\alpha_{1}C_{1}(1,t)&  \alpha_{2}C_{1}(2,t)  &\cdots &\alpha_{K}C_{1}(K,t)\\
    & \alpha^{2}_{1}C_{1}(1,t) & \alpha^{2}_{2}C_{1}(2,t) &\cdots &\alpha^{2}_{K}C_{1}(K,t) \\
    &\vdots &\vdots &\vdots\\
     &\alpha^{N}_{1}C_{1}(1,t) &\alpha^{N}_{2} C_{1}(2,t) &\cdots &\alpha^{N}_{K}C_{1}(K,t)
    \end{bmatrix}
    \begin{bmatrix} 
    1\\1\\\vdots\\1
    \end{bmatrix}=
    \begin{bmatrix} 
    0\\0\\0\\\vdots\\0
    \end{bmatrix}
\end{equation}
With $t\not\in\{1,2,\cdots,K\}$, it is guaranteed that $C_{1}(k,t)=k-t\neq 0$, $k\in\{1,\cdots,K\}$. We have
\begin{equation}\label{vand1}
\setlength{\arraycolsep}{1.0pt}
    \begin{bmatrix} & \frac{C_{0}(1,t)}{C_{1}(1,t)} & \frac{C_{0}(2,t)}{C_{1}(2,t)}&\cdots &\frac{C_{0}(K,t)}{C_{1}(K,t)}\\
    & 1 & 1 &\cdots &1 \\
    &\alpha_{1}&  \alpha_{2}  &\cdots &\alpha_{K}\\
    
    &\vdots &\vdots &\vdots\\
     &\alpha^{N}_{1} &\alpha^{N}_{2}  &\cdots &\alpha^{N}_{K}
    \end{bmatrix}
    \begin{bmatrix} 
    \alpha_{1}C_{1}(1,t)\\\alpha_{2}C_{1}(2,t)\\\vdots\\\alpha_{K}C_{1}(K,t)
    \end{bmatrix}=
    \begin{bmatrix} 
    0\\0\\0\\\vdots\\0
    \end{bmatrix}
\end{equation}
Third, we remove the first line and denote $F\theta^{*}=\mathbf{0}^{*}$, where
\begin{align*}\label{prove1}
\setlength{\arraycolsep}{1.0pt}
 F = \begin{bmatrix} 
    & 1 & 1 &\cdots &1 \\
    &\alpha_{1}&  \alpha_{2}  &\cdots &\alpha_{K}\\
    &\vdots &\vdots &\vdots\\
     &\alpha^{N}_{1} &\alpha^{N}_{2}  &\cdots &\alpha^{N}_{K}
    \end{bmatrix},
\theta^{*} =  \begin{bmatrix} 
    \alpha_{1}C_{1}(1,t)\\\alpha_{2}C_{1}(2,t)\\\vdots\\\alpha_{K}C_{1}(K,t)
    \end{bmatrix},
    \mathbf{0}^{*}=
    \begin{bmatrix} 
    0\\0\\0\\\vdots\\0
    \end{bmatrix}
\end{align*}
Since the coefficient matrix $F$ is a Vandermonde matrix, the solution $\theta^{*}$ is a zero solution. Because $C_{1}(i,t)\neq 0 $, $\{\alpha_{1},\cdots,\alpha_{K}\}$ are all zeros. Thus, it is a clear contradiction and the individual column vectors $\{A_{(1)},\cdots,A_{(K)}\}$ are linearly independent.
\end{proof}

\section{Experimental Settings}\label{app::experimentsetting}
{\bf Dataset.}~~

For the sake of fairness, we select benchmark datasets that are commonly used in related work. 
For the homophilic graph, we select three classic citation graphs: Cora, CiteSeer, PubMed \cite{Cora}. For the heterophilic graph, we select four datasets including the Webpage graphs Texas , Cornell and Wisconsin from WebKB3 \footnote{http://www.cs.cmu.edu/afs/cs.cmu.edu/project/theo-11/www/wwkb}, and the Actor co-occurrence graph \cite{Gemo}. Besides, we also select Penn94 from Facebook 100 \footnote{https://archive.org/details/oxford-2005-facebook-matrix} as a representative dataset of the social network\cite{LINKX}.  We give the statistics for these datasets in Table \ref{tab::datasets}, in which we use edge homophily ratio $ h(G) = \frac{\left|\left\{ (u, v) : (u, v) \in \mathcal{E} \wedge y_{u} = y_{v} \right\}\right|}{|\mathcal{E}|}$ to measure the degree of homophily \cite{H2GCN}.

\begin{table}[!ht]
\centering
\caption{Dataset statistics.}\label{tab::datasets}
\begin{center}
\begin{small}
\resizebox{0.45\textwidth}{!}
{\begin{tabular}{l c c c c c c c c c c}
\toprule
Datasets   &Cora &Citeseer &Pubmed  &Actor &Texas &Cornell & Wisconsin  & Penn94\\
\midrule
 Nodes    &2708 &3327 &19,717  &7600 &183 &183  &251    &41,554 \\
 Edges   & 5278 &4676 &44,327 &26,752 &295 &280 &466  &1,362,229\\
 Features   &1433 &3703 &500  &931 &1703 &1703 &1,703  &5 \\
Classes   &7 &6 &3  &5 &5 &5  &5 &2 \\
$h(G)$  &0.81 &0.74 &0.80 &0.22 &0.11 &0.30 &0.21  &0.47\\ 
\bottomrule
\end{tabular}}

\end{small}
\end{center}
\vskip -0.1in
\end{table}

{\bf Train/validation/test split.}~~

For semi-supervised node classification in section {5.1}, we split the dataset with random seeds according to \cite{Chebyshev2}. In detail, we select 20 nodes of each class on three homophilic datasets (Cora, Citeser, and Pubmed) for training, 500 nodes for validation, and 1000 nodes for testing. For the five heterophilic datasets, we apply the sparse splitting where the training/validation/test
sets account for 2.5\%/2.5\%/95\%.

For node classification, there are two common split ways: fixed seed splits and random seed splits. The former uses the fixed split given in \cite{Global-hete}, which is commonly used in heterophilic GNNs. The latter, proposed by \cite{GPRGNN}, uses random seeds for random split and is widely used in spectral GNNs, PDE GNNs, traditional methods, etc. To avoid unfair comparison, we evaluate the performance of PCNet under these two different settings. 
For the node classification with polynomial-based methods in section {5.2}, we adopt 10 random seeds as \cite{GPRGNN,JacobiConv} to randomly split the dataset into train/validation/test sets with a ratio of 60\%/20\%/20\%. In section {5.3},
we use the given split in  \cite{Global-hete} to randomly split the dataset into training/validation/test sets at a ratio of  48\%/32\%/20\%.

{\bf Training process.}~~

We utilize Adam optimizer to optimize models and set the maximum number of iterations to 1000. In addition, we use the early stopping strategy to end training if the validation scores on the real-world dataset don't increase over time. For sections {5.1} and {5.2}, we set the early-stop epoch threshold to 200, in line with \cite{Chebyshev2,JacobiConv,BernNet}. For section {5.3}, we use the same settings as GloGNN++ \cite{Global-hete} and set the early-stop epoch threshold as a hyperparameter in the range $\{40, 200, 300\}$. For dataset Penn94, we use AdamW as the optimizer following \cite{Global-hete}.

{\bf Computing infrastructure.}~~ 

We leverage Pytorch Geometric and Pytorch for model development. All experiments are conducted on the same machine with the Intel(R) Core(TM) i7-8700 3.20GHz CPU, two GeForce
GTX 1080 Ti GPUs and 64GB RAM. The implementation of PCNet will be public available upon acceptance.

\end{document}